\documentclass[letterpaper, 10 pt, conference]{ieeeconf}  

\IEEEoverridecommandlockouts    
\usepackage{vcell}
\usepackage{booktabs}
\usepackage{graphicx}
\usepackage[colorlinks,linkcolor=red,anchorcolor=blue,citecolor=green]{hyperref}
\usepackage{cite}
\usepackage{array}
\usepackage{multirow}

\usepackage{graphicx} 
\usepackage{amsmath} 
\usepackage{amssymb}  

\usepackage[capitalize]{cleveref}
\crefname{section}{Sec.}{Secs.}
\Crefname{section}{Section}{Sections}
\Crefname{table}{Table}{Tables}
\crefname{table}{Tab.}{Tabs.}
\usepackage[table,xcdraw]{xcolor}

\title{\LARGE \bf
\ Tracking Any Point with Frame-Event Fusion Network at High \ Frame Rate
}

\author{Jiaxiong Liu, Bo Wang, Zhen Tan, Jinpu Zhang, Hui Shen$^*$, Dewen Hu$^*$ 
\thanks{J. Liu, B. Wang, Z. Tan, J. Zhang, H. Shen, D. Hu are with the College of Intelligence Science and Technology, National University of Defense Technology, China. \{liujiaxiong21, wb, tanzhen1996 \}@nudt.edu.cn}%
\thanks{* indicates corresponding authors:  H. Shen (shenhui@nudt.edu.cn) and D. Hu (dwhu@nudt.edu.cn)}
}

\begin{document}

\maketitle
\thispagestyle{empty}
\pagestyle{empty}

\begin{abstract}
Tracking any point based on image frames is constrained by frame rates, leading to instability in high-speed scenarios and limited generalization in real-world applications.
To overcome these limitations, we propose an image-event fusion point tracker, FE-TAP, which combines the contextual information from image frames with the high temporal resolution of events, achieving high frame rate and robust point tracking under various challenging conditions. Specifically, we designed an Evolution Fusion module (EvoFusion) to model the image generation process guided by events. This module can effectively integrate valuable information from both modalities operating at different frequencies. To achieve smoother point trajectories, we employed a transformer-based refinement strategy that updates the point's trajectories and features iteratively. Extensive experiments demonstrate that our method outperforms state-of-the-art approaches, particularly improving expected feature age by 24$\%$ on EDS datasets. Finally, we qualitatively validated the robustness of our algorithm in real driving scenarios using our custom-designed high-resolution image-event synchronization device. Our source code will be released at https://github.com/ljx1002/FE-TAP.
\end{abstract}

\section{INTRODUCTION}

Establishing point correspondences is a fundamental vision task and has been extensively applied across various domains, including autonomous driving and simultaneous localization and mapping (SLAM).
Despite significant advances in the performance of point trackers based on traditional cameras in recent years \cite{conf_nips_DoerschGMRSACZY22,conf_iccv_wang23,conf_iccv_carl23,conf_eccv_HarleyFF22,journals_corr_abs_2307_07635,journals_corr_abs_2403_14548}, their accuracy is still limited in extreme scenarios, such as high-speed motion and low-light conditions, due to inherent hardware constraints.

\begin{figure}[t]
    \centering
    \includegraphics[width=1\linewidth]{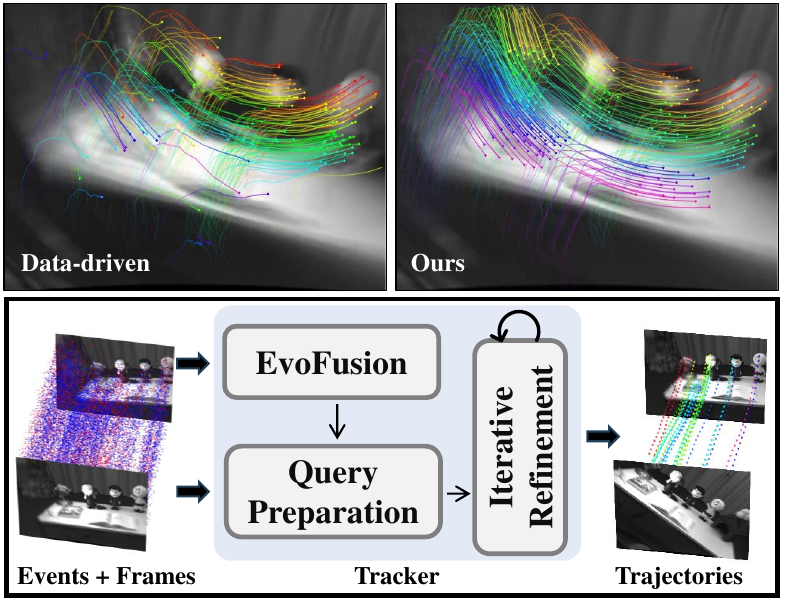}
    \caption{Comparison of tracking performance in high-speed motion scenarios: Our method (top right), integrating image and event data, vs. Data-driven methods (top left),  which rely on the first image frame and event data.}
    \label{fig:first}
    \vspace{-0.3cm}
\end{figure}

Event cameras, inspired by the principles of the human retina, can overcome these limitations. 
By independently sensing logarithmic changes in brightness at each pixel, event cameras output event streams with microsecond-level temporal resolution, offering advantages such as high dynamic range and low power consumption. Currently, event-based point trackers have shown promising results in high-speed and HDR scenes~\cite{journals_ijcv_GehrigRGS20,conf_cvpr_MessikommerFG023}. The majority of event-based trackers are built upon classical models~\cite{journals_ijcv_GehrigRGS20,conf_icra_ZhuAD17,conf_bmvc_AlzugarayC20}, which are significantly impacted by the quality of event data. As event noise increases, tracking performance rapidly deteriorates. Data-driven~\cite{conf_cvpr_MessikommerFG023} proposed the first neural network-based point tracker, which markedly improved tracking performance without requiring parameter adjustments for different scenes. Nonetheless, due to the lack of intensity and detailed texture information in event data, achieving robust tracking in complex environments remains a significant challenge. 

Therefore, we aim to fuse low-frequency but texture-rich image frames with high-frequency event data to enable the tracking of any points in various motion scenarios. To attain our objective, two challenges need to be addressed: (i) The measurement rate of aggregated events is significantly higher than that of image frames. Direct fusion of low-frequency images with high-frequency events can lead to spatial misalignment, negatively impacting downstream tasks. Although several methods combining images and events have been proposed in fields such as feature point detection~\cite{conf_icra_WangCYYXY24,journals_ral_WangYYYX24}, line segment detection~\cite{journals_pami_yu23}, and object tracking~\cite{conf_cvpr_zhang23, wang2023visevent,zhou2023rgb}, these approaches either have their output frame rates restricted by the image frame rates or rely on complex temporal alignment strategies. (ii) Effectively leveraging both modalities to achieve any point tracking across different motion scenarios presents another challenge. To the best of our knowledge, no existing work has utilized image and event to achieve any point tracking.

To tackle these deficiencies, we propose the first data-driven tracker (FE-TAP) that integrates both image frames and event data to track any point. 
Specifically, we first propose an evolution fusion module (EvoFusion) to fuse events and image frames with different frame rates. In contrast to previous approaches that rely on time alignment modules, which are difficult to model due to the requirement for accurate camera motion and depth information, often resulting in substantial errors, EvoFusion offers a new perspective. 
Our module fuses images with all subsequent events by utilizing a well-designed convolutional network to learn the gradual evolution of images under the influence of events. This process generates the latest image-like information, effectively leveraging the strengths of both modalities. Our module can rely on event information to restore image features when the input image is blurry, resulting in robust fused features.

Then we introduce a designed transformer-based module to capture the spatio-temporal relationships between target points during trajectory optimization. This model operates in a sliding window fashion on a two-dimensional representation of a token. The transformer uses attention mechanisms to consider each track in its entirety within a window and exchange information between tracks, resulting in smoother trajectories. To better adapt to the image-event fusion tracking task, we also encoded the event accumulation time for each fused feature and incorporated it into the token. Additionally, by optimizing the trajectories within a sliding window, our algorithm inherently possesses a degree of occlusion robustness. Our tracker outperforms existing approaches method by 5$\%$ on the EC Dataset~\cite{jour_ijrr_mueggler_2017event} and by 24$\%$ on the EDS dataset~\cite{conf_cvpr_Hidalgo_Carrio022}.

The main contributions are listed as follows:
\begin{itemize}
\item We propose the first data-driven tracker that fuses image frames and event data to track any point.
\item We design an Evolution Fusion module to combine frames and events at different frequencies, enabling stable performance of our tracker in extreme scenarios.
\item We introduce a transformer-based module that captures the spatio-temporal relationships between target points to optimize their trajectories within a sliding window.
\item The superior performance of our method is validated on public datasets and further confirmed with real driving data that captured by our custom-designed high-resolution image-event synchronization device.

\end{itemize}

\section{RELATED WORK}

\subsection{Frame-Based methods}

The problem of tracking any points was recently introduced in TAP-Vid~\cite{conf_nips_DoerschGMRSACZY22}, which focuses on estimating the motion of any points over time. 
PIP~\cite{conf_eccv_HarleyFF22} revisits the classic particle video problem by leveraging entire image sequences to query point trajectories. This method effectively addresses occlusions in intermediate frames by utilizing the rich contextual relationships between target points. CoTracker~\cite{journals_corr_abs_2307_07635} considers the significant spatial correlations between target points due to rigid connections in the physical world. This approach improves tracking performance by jointly tracking all target points across multiple frames and introduces a sliding window design that enables online tracking. Another area related to any point tracking is optical flow estimation~\cite{conf_eccv_TeedD20,conf_iccv_alexey15,journals_ijcv_WangZLYSLH24,Shi_2023_CVPR}, which involves estimating dense pixel correspondences between consecutive frames. These methods face difficulties in achieving long-term point tracking. Despite the notable achievements of image-based point trackers in recent years, inherent hardware limitations of standard cameras prevent them from effectively addressing any point tracking tasks in high-speed or low-light scenarios. Additionally, these systems face challenges related to the trade-off between bandwidth and latency.

\subsection{Event-Based methods}

In recent years, using novel event cameras to track points in challenging scenarios has gained significant popularity. Early event-based feature point trackers were developed based on classical models. For example, research~\cite{conf_icra_ZhuAD17} processes event streams as point clouds and uses the ICP algorithm to estimate feature point trajectories. HASTE~\cite{conf_bmvc_AlzugarayC20} updates feature point trajectories on a per-event basis by hypothesizing $11$ possible motion patterns for the feature points and matching templates to identify the most likely motion outcome. EKLT~\cite{journals_ijcv_GehrigRGS20} uses grayscale images as templates and matches them with brightness increment images derived from event streams to achieve feature point tracking. Recently, ~\cite{conf_cvpr_MessikommerFG023} introduced the first neural network-based model for feature point tracking with event cameras, significantly enhancing performance in challenging environments.

However, these algorithms rely solely on events or the initial image frame and events, resulting in poor performance on complex datasets. The high noise levels in event data and the lack of detailed texture information make it difficult to maintain robust tracking in intricate environments.

In a similar direction to feature point tracking, several works have proposed various feature point detectors for event cameras.~\cite{conf_cvpr_ChiberrePSL21,journals_ral_AlzugarayC18,conf_icra_WangCYYXY24,journals_ral_WangYYYX24,conf_iros_LiSZLL19,journals_corr_abs_2209_10385}. These methods leverage the strong spatio-temporal relationships between event features to directly track feature points. For instance, FE-DeTr~\cite{conf_icra_WangCYYXY24} combines event streams and image frames, using a self-supervised strategy for keypoint detection, and then tracks feature points by utilizing the spatio-temporal relationships between them. However, these methods are unable to track arbitrarily specified points.

Inspired by these advances, we leverage neural networks to fuse the low-frequency but texture-rich image frames with the sparse yet high-frequency event streams at the feature level. We then optimize the target point trajectories using a transformer-based module, enabling achieving high-frequency and stable tracking of any point.

\section{METHOD}
\begin{figure*}[t!]
    \vspace{0.15cm}
    \centering
    \includegraphics[width=1\linewidth]{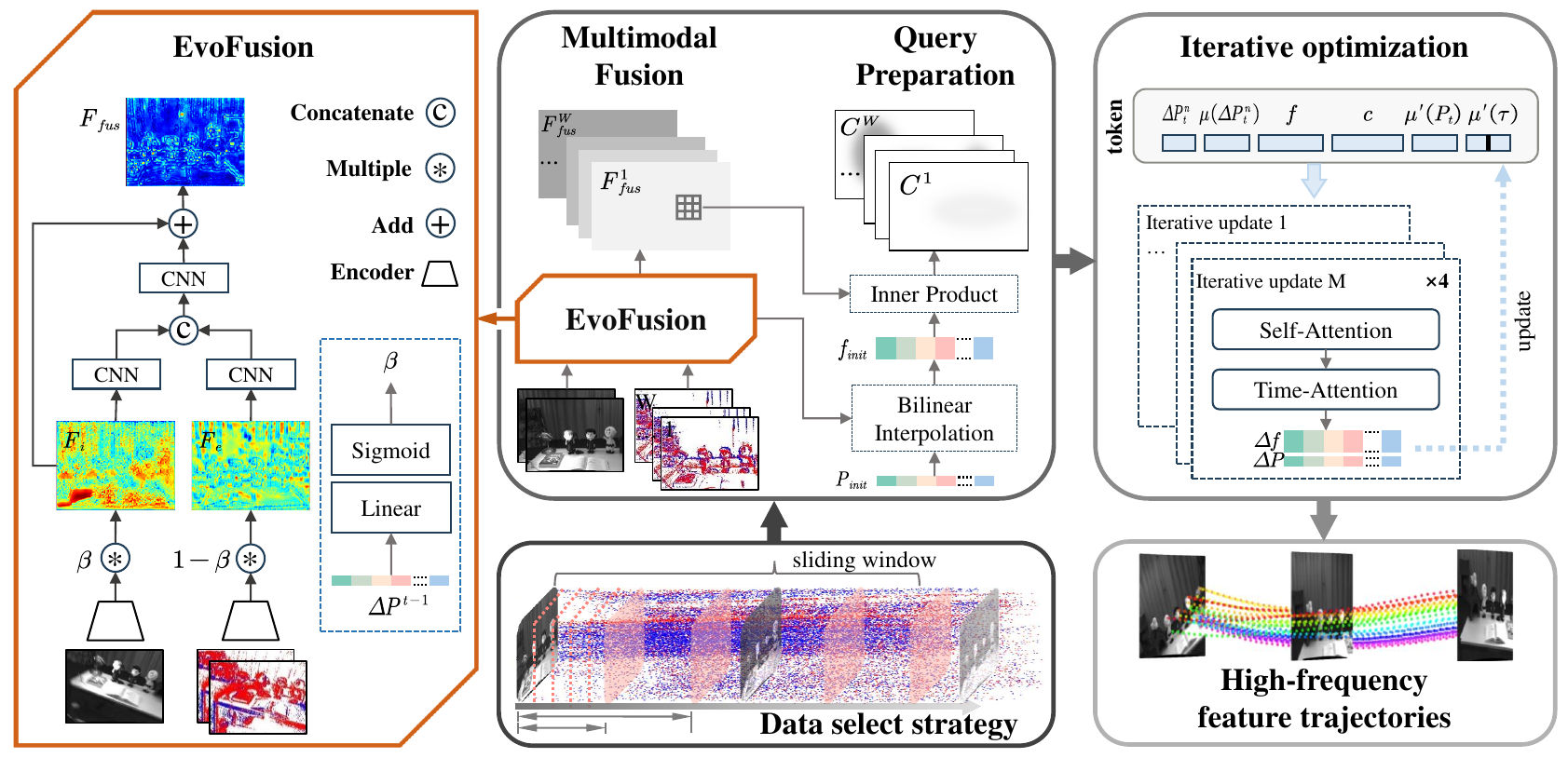}
    \caption{The overview of FE-TAP. EvoFusion module fuses image and event data with different frame rates using an appropriate data selection strategy. The query preparation module computes cost volumes based on the fused feature maps. The iterative update module takes these elements as input and optimizes all point query trajectories in parallel within a sliding window, producing high-frequency point tracks.}
    \vspace{-0.35cm}
    \label{fig_structure}
\end{figure*}

The overall architecture of our network is shown in  \cref{fig_structure}. First, we use the Evolution Fusion module (EvoFusion) to fuse image frames and event representations (see \cref{Event representation}) to produce high-frequency fused features map $F_{fus}$. 
Next, query content features $f_{init}$ and a correlation volume $C^{w}$ are computed based on the fused feature map and the query point position $P_{init}$, which represents the initial location of all target points to be tracked over time. 
Leveraging the strong contextual understanding and the efficient parallel processing capabilities of the transformer, point trajectories are iteratively refined in a sliding window fashion, enabling robust long-term point tracking.
Notably, due to the sliding window trajectory optimization, our module exhibits a certain level of occlusion robustness, even without explicitly accounting for occlusions. The entire process operates at a high temporal resolution and is not constrained by the frame rate.

\subsection{Event representation} 
\label{Event representation}
To use asynchronous event streams as input to a neural network, we must first convert the event stream into a tensor-like matrix while retaining as much useful information as possible. We adopt a representation method similar to Stacking Based on Time (SBT)~\cite{conf_cvpr_WangIHY19}. For the selected event stream $E = \{e_i\}_{i=1}^{N}$ between timesteps $t_{start}$ and $t_{end}$, each event $e_i$ contains pixel coordinates $x_i, y_i$, a timestamp in microseconds $t_i$, and the polarity $p_i \in \{-1, 1\}$ indicating the brightness change. The event stream is then divided into $B$ bins based on time, with the pixel values in each bin assigned the normalized timestamp of the most recent event, as shown in the following equation:
\begin{gather}
    S(x,y,t) = \max_{\substack{t_i^* \in [t, t+1)}} \left( k(x - x_i) \cdot k(y - y_i) \cdot t_i^* \right)   ,\label{eq_SBT_1}\\
    t_i^* = \frac{t_i - t_{start}}{t_{end} - t_{start}}(B-1),\label{eq_SBT_2}\\
    k(a) = \max(0, 1-|a|).\label{eq_SBT_3}
\end{gather} 
Here, $x$, $y$, and $t$ represent the x-y-time dimensions of the event representation S. Considering the polarity of events, the output dimension of event representation is $(X, Y, 2B)$.

\subsection{EvoFusion}
EvoFusion is designed to integrate image frames and event representations with varying frequencies, extracting complementary information to generate high-frequency fused features. The key challenges that EvoFusion addresses are: (i) Cross-frame-rate alignment, where event representations have a much higher frequency than image frames, potentially leading to spatial misalignment and blurred features if combined directly; and (ii) Maintaining robust tracking performance in both static and high-motion scenarios by effectively leveraging the advantages of both modalities.

To address the first challenge, we employ a data selection strategy that integrates images and events, using a network to model the image generation process. This allows for the fusion of image frames and events at different frequencies. 
It is sufficient to reconstruct the absolute brightness of the environment at any given moment after the image frame was captured by using the frame and subsequent event stream (despite events encoding logarithmic brightness changes). Thus, we can avoid the errors introduced by complex temporal alignment modules, providing a simpler and more effective method for fusing images and events at different frequencies.

Here, we employ two convolutional encoders, built on Feature Pyramid Networks (FPN)~\cite{conf_cvpr_lin17}, with identical architectures but without shared weights to extract features from images and event representations. 
The encoders transform the input event representation from a size of ($H\times W \times2B$) or the image frame from a size of ($H \times W \times 3$) to a feature map of size ($\frac{H}{S} \times \frac{W}{S} \times C$), where $C=128$ is the feature map's channel size, and $S=4$ is the downsampling factor.

Both image frames and event data play crucial roles in point-tracking tasks. To adaptively extract and integrate complementary information from both modalities, we designed a feature fusion module to address the second challenge, as illustrated in  \cref{fig_structure}. The process is formulated as follows:
\begin{gather}
    F_e^t = \text{ReLU}(\text{Conv}(F_e^t)), \\
    F_i^t = \text{ReLU}(\text{Conv}(F_i^t)), \\
    F_{fus}^t = \text{ReLU}(\text{Conv}( \beta \cdot F_i^t) + (1-\beta) \cdot F_e^t) + \tilde{F_i^t}), \\
    \beta = \text{Sigmoid} ( \text{Linear} ({\Delta P}^{t-1})).
\end{gather}
Here, $\beta$ and $(1 - \beta)$ represent the weights of the image feature maps $F_i^t$ and event feature maps $F_e^t$, respectively. These weights are determined by a linear network module based on the average optical flow magnitude $\Delta P$, from the previous moment. The fused feature map is denoted as $F_{fus}^j$.

\subsection{Query Preparation}
We introduce a sliding window approach to optimize point trajectories. Within each window, a transformer-based module is used to fully exploit the temporal correlations of individual point trajectories and the spatial correlations between different points at each moment, resulting in more accurate point tracking. So, we first need to prepare the tokens for the transformer-based module.

At each moment, every tracking point is assigned to a point query. The point query associated with the $n$-th tracking point in the $t$-th moment is tasked with identifying the most matching point of its content feature in the $t$-th moment. To obtain an accurate content feature vector of the point query, we perform bilinear sampling on the image feature map based on the initial position of the point query. The initial content feature for a point query is extracted from the image feature map because image information is not affected by the relative motion between the camera and the external environment. For a fixed-size sliding window of length $W$, we initialize by replicating the initial content feature along the time dimension. The target trajectory's positional is also initialized in the same manner. 

To assess the accuracy of the current trajectory, the correlation volume between the content feature vector and the feature vectors extracted from each pixel location in the fused feature map must also be calculated. Specifically, the correlation vectors in the correlation volume are formed by stacking the inner products between the content feature vector and multiple fused feature vectors surrounding the predicted position of the point query. To capture multi-scale information in the correlation volume, average pooling is applied to the fused feature maps, which are then used to derive the multi-scale correlation volume $C^{w}$. For non-integer positions or those near the border, bilinear interpolation and zero-padding are employed for sampling. 

\subsection{Iterative Refinement} 
In the trajectory optimization stage, a sliding window with a time step $T_{step}$ is used, where $T_{step} = 1$ corresponds to real-time operation and $T_{step} < W$. This setup allows the transformer-based module to update the point query trajectories and its content feature vectors within the window. The input token consists of its displacement, displacement encoding, content feature vector, correlation vector, time encoding, and positional encoding:
\begin{equation}
    \label{eq_iterative_update}
G_t^n = \left( \hat{P}_t^{n} - \hat{P}_1^{n}, f_t^n, C_t^n, \mu(\hat{P}_t^{n} - \hat{P}_1^{n}) , \mu'(\hat{P}_t^{n}, \mathcal{T})\right).
\end{equation}
Here, $\hat{P}_t^{n}$ denotes the predicted position of the $n$-th point query at time $t$, with $t = 1$ indicating the initial time within the sliding window. The function $\mu$ denotes a sinusoidal positional encoding, while $\mu'$ encodes both the initial position of the point query and the time information $\mathcal{T}$, with parameters fine-tuned based on the final results. To enhance the effectiveness of the transformer-based iterative module in utilizing fused feature maps, the accumulated event duration $\mathcal{T}$ for each fused feature is encoded and incorporated into the tokens, thereby accelerating the convergence process. 
The module outputs $\Delta P$ and $\Delta f$, which represent changes in point query position and content feature vectors, respectively. To achieve more precise tracking, multiple optimizations are performed on the point trajectories within each window. Importantly, $\Delta f$ only influences subsequent iterations within the current window and does not modify the query point's content feature vector template in future windows, thus preventing the accumulation of errors.

\section{Experiments}
Our model is trained on the synthetic Multiflow dataset, which provides ground-truth optical flow~\cite{jour_pami_gehrig24}. For evaluation, we test on two public real-world datasets: the Event Camera Dataset (EC)~\cite{jour_ijrr_mueggler_2017event} and the Event-aided Direct Sparse Odometry Dataset (EDS)~\cite{conf_cvpr_Hidalgo_Carrio022}. We visualize our tracking results on several representative scenes from the test datasets to demonstrate the effectiveness of our method, as shown in  \cref{fig:result_tracking}. Finally, we used our custom-designed high-resolution image-event synchronization device to qualitatively test our algorithm in real driving scenarios with moving objects.

\begin{table*}[t!]
\vspace{0.15cm}
\centering
\caption{Performance comparison of trackers on the EDS and EC dataset, with the best results in \textbf{bold}, second-best results \underline{underlined}, and * indicating that the ground truth of the dataset is not fully accurate.}
\label{tab:result_all}
\vspace{-0.3cm}  
\renewcommand{\arraystretch}{1.2}  
\begin{tabular}{>{\centering\arraybackslash}m{2.4cm}  
    >{\centering\arraybackslash}m{0.9cm}  
    >{\centering\arraybackslash}m{0.9cm}  
    >{\centering\arraybackslash}m{0.9cm}  
    >{\centering\arraybackslash}m{1.4cm}  
    >{\centering\arraybackslash}m{1.0cm}  
    c
    >{\centering\arraybackslash}m{0.9cm}  
    >{\centering\arraybackslash}m{0.9cm}  
    >{\centering\arraybackslash}m{0.9cm}   
    >{\centering\arraybackslash}m{1.4cm}  
    >{\centering\arraybackslash}m{1.0cm} }
\hline
\multirow{3}{*}{\centering Sequence} & \multicolumn{5}{c}{Feature Age $\uparrow$}   &      & \multicolumn{5}{c}{Expected Feature Age $\uparrow$}                         \\ \cline{2-6}  \cline{8-12} 
&  ICP & HASTE &  EKLT &  Data-driven (SOTA)   &   FE-TAP (OURS) &          &   ICP & HASTE &  EKLT &  Data-driven (SOTA)   &   FE-TAP (OURS)  \\  \hline
shapes translation   &  0.307      &    0.589       &      \underline{0.839}    &      0.817      &   \textbf{0.931}     & &  0.306      &    0.564       &      0.740    &      \underline{0.810}     &   \textbf{0.929}  \\
shapes rotation  &     0.341            &          0.613       &     \underline{0.833 } &          0.791       &     \textbf{0.815}   &  &       0.339       &     0.582            &          \underline{0.806}       &     0.786            &          \textbf{0.813}       \\
shapes 6DOF  &        0.169         &              
     0.133   &        0.817         &   \textbf{0.917} &    \underline{0.879} &    &         0.129   &        0.043         &              0.696   &   \textbf{0.899 }    &    \underline{0.860}   \\
boxes translation  &      0.268           &        0.382         &      0.682           &        \textbf{0.863}&      \underline{0.731}        & &      0.261         &      0.368           &        0.644         & \textbf{0.858}    &        \underline{0.728}         \\
boxes rotation  &      0.191           &           0.492      & \textbf{0.883 }      &           0.640      &      \underline{0.862}     &   &       0.188      &      0.447           &        \textbf{0.865}&      0.637           &           \underline{0.861}      \\
\hline
EC Avg      &      0.256           &             0.442    &      \underline{0.811}           &             0.805    & \textbf{0.844 }      &    &        0.245    &      0.427           &             0.775    &      \underline{0.798}           &             \textbf{0.838}    \\
\hline
Peanuts Light     &      0.050           &        0.086         &      0.284           &        \underline{0.446}         &\textbf{0.549}        &  &     0.044         &      0.076           &        0.260         &      \underline{0.423}           &        \textbf{0.517}         \\
Rocket Earth Light*     &      0.103           &      0.162           &      0.425           &      \textbf{0.654}  &      \underline{0.538}     &     &      0.045           &      0.085           &      0.175           &\textbf{0.296}        &      \underline{0.246}           \\
Ziggy In The Arena     &       0.043          &      0.082           &       0.419          &      \underline{0.729}           & \textbf{0.849}       &     &    0.039           &       0.057          &      0.231           &       \underline{0.727}          &      \textbf{0.844}      \\
Peanuts Running     &       0.043         &         0.054         &       0.171         &         \underline{0.482}         & \textbf{0.769}       & &       0.028         &       0.033         &         0.153         &       \underline{0.455}         &         \textbf{0.749}        \\
\hline
EDS Avg   &  0.060  &   0.096   &  0.325   &     \underline{0.577}  &   \textbf{0.676 } & & 0.060 & 0.161 & 0.325 &   \underline{0.475}  &    \textbf{0.589 }                         \\
\hline
\end{tabular}
\vspace{-0.2cm}  
\end{table*}

\subsection{Implementation Details}
\label{Implementation Details}
The model is supervised by calculating the $\mathcal{L}_1$ distance between the estimated trajectory and the ground-truth trajectory. This supervision is performed across multiple iterative updates, applying exponentially increasing weights, as in RAFT~\cite{conf_eccv_TeedD20}. 
Specifically, the RAFT loss function is applied within each sliding window, and the losses are accumulated across all windows to ensure overall optimization.

We configure our system with the following parameters: $B = 5$ for event representation, $W = 16$ for the length of the sliding window, $T_{step} = 8$ for the time step of the sliding window, $M = 4$ for the number of iterative updates and Adam-W optimizer was used with an initial learning rate of 0.0005, employing a dynamic adjustment strategy that increased the rate initially and then gradually decreased it. The total number of training steps was set to 150,000.

\subsection{Datasets and Metrics}

The EC dataset is recorded using the DAVIS240C camera~\cite{jour_ssc_brandli14}, which provides 240 $\times$ 180 resolution image frames at 24Hz along with corresponding event data. 
Ground-truth camera poses are available at 200Hz. The EDS dataset is captured using a setup consisting of an RGB camera and an event camera with the same resolution. 
This configuration produces higher resolution image frames and event data (640 $\times$ 480 pixels). Similar to the EC dataset, the EDS dataset includes ground-truth camera poses at 150Hz. The ground truth trajectories for both datasets are obtained by calculating the 3D coordinates of target points and projecting them to 2D based on the camera positions.

For evaluation, two widely used metrics are employed: Feature Age (FA) and Expected Feature Age (EFA). FA measures the percentage of a target point's ground-truth lifespan during which it is tracked within a certain pixel error threshold. The EFA metric takes into account the impact of points that were lost at the beginning of tracking. For more details on these performance metrics, please refer to~\cite{conf_cvpr_MessikommerFG023}.

\subsection{Result Comparisons}
\begin{figure}[t]
    \centering
    \includegraphics[width=1\linewidth]{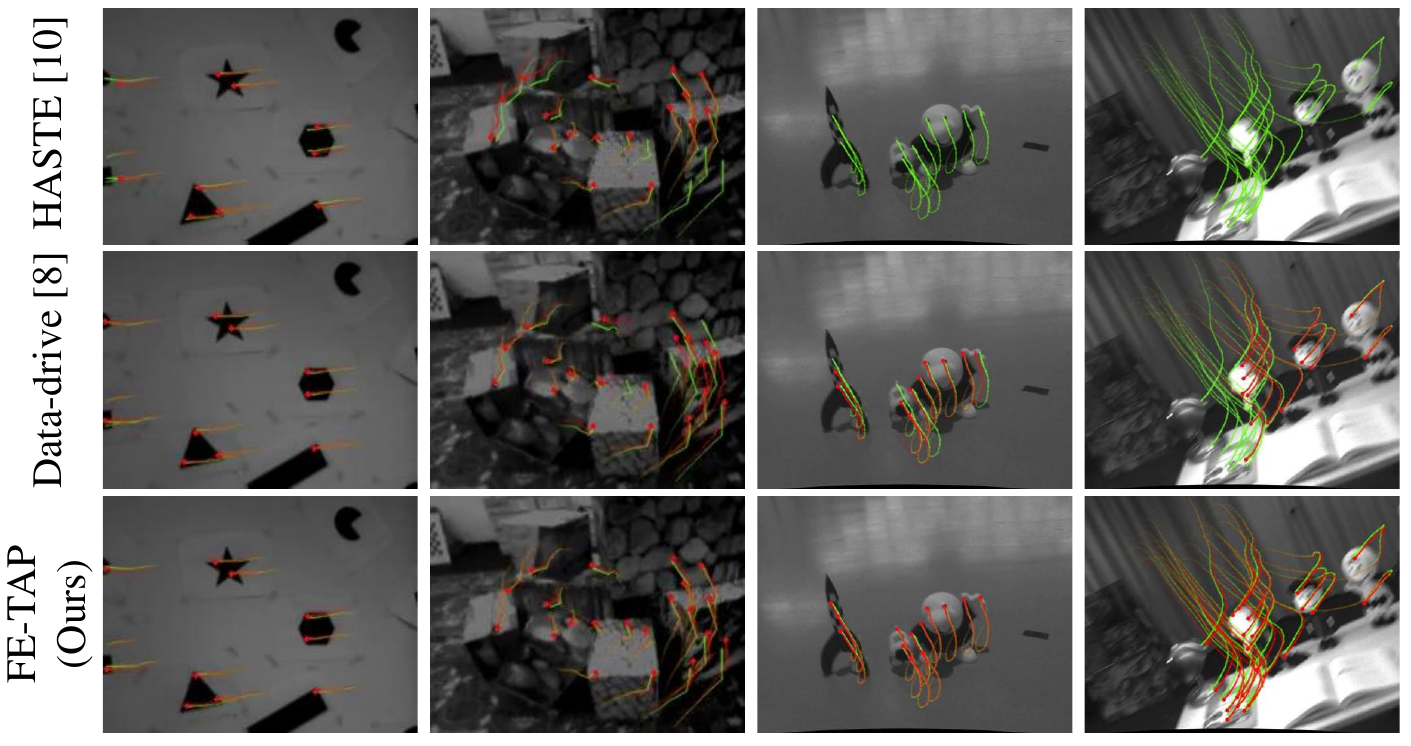}
    \caption{Qualitative tracking predictions(red) and ground truth tracks(green) for EC dataset (1st, 2nd col) and EDS dataset (3rd, 4th col). We discard predicted trajectories if they deviate significantly from the ground truth trajectory.}
    \vspace{-0.3cm}
    \label{fig:result_tracking}
\end{figure}
To validate the effectiveness of our method, we selected several representative high-frame-rate point tracking methods and SOTA approaches for comparison. These include (1) ICP tracker~\cite{kueng2016low}, which uses grayscale images as templates and subsequently relies on event streams for feature tracking, commonly used in event-based visual odometry; (2) HASTE~\cite{conf_bmvc_AlzugarayC20}, a purely event-based tracker; (3) EKLT~\cite{journals_ijcv_GehrigRGS20}, which extracts template patches from the first frame and tracks using the event stream; and (4) Data-driven~\cite{conf_cvpr_MessikommerFG023}, the current SOTA for event-based tracking, a learning-based method that utilizes the initial image frame and subsequent event streams for tracking. Each method was fine-tuned for specific datasets to achieve optimal performance, while our method was applied directly after training on the synthetic dataset, without additional scene-specific parameter tuning.

As shown in \cref{tab:result_all}, our proposed FE-TAP method outperformed the baselines across both datasets, achieving the best results in terms of FA and EFA.
Specifically, our method improved EFA by 5$\%$ and 24$\%$ compared to the SOTA method. The EC dataset, with relatively simple environments and motion, posed fewer challenges for tracking, leading to impressive results from EKLT, Data-driven, and our method. Despite employing a downsampling operation that inherently reduces tracking precision, our method still achieved superior results, as demonstrated in the first and second cols of  \cref{fig:result_tracking}.

The EDS dataset presented greater challenges due to its inclusion of more complex and rapid camera movements, intricate background information, and higher levels of noise. nevertheless, our method demonstrated significant improvements over existing methods in both FA and EFA, as indicated in the last two cols of  \cref{fig:result_tracking}. This success can be attributed to our fusion module, which effectively leverages the complementary strengths of image and event data. In these high-resolution datasets, images can capture more detailed texture information, which is crucial for target point tracking. Additionally, the stable information from the images helps distinguish noisy events, resulting in more reliable and robust tracking. The significant performance gains validate that our image-event fusion method effectively handles more challenging scenes with complex 3D structures, varying motion conditions, and noise patterns.
It is worth noting that in the EDS dataset, the Rocket Earth Light sequence contains occlusions that result in inaccurate ground truth. When this sequence is excluded, our method outperforms existing methods in EFA by up to 31.5$\%$.

\begin{figure}[t!]
    \vspace{0.15cm}
    \centering
    \includegraphics[width=1\linewidth]{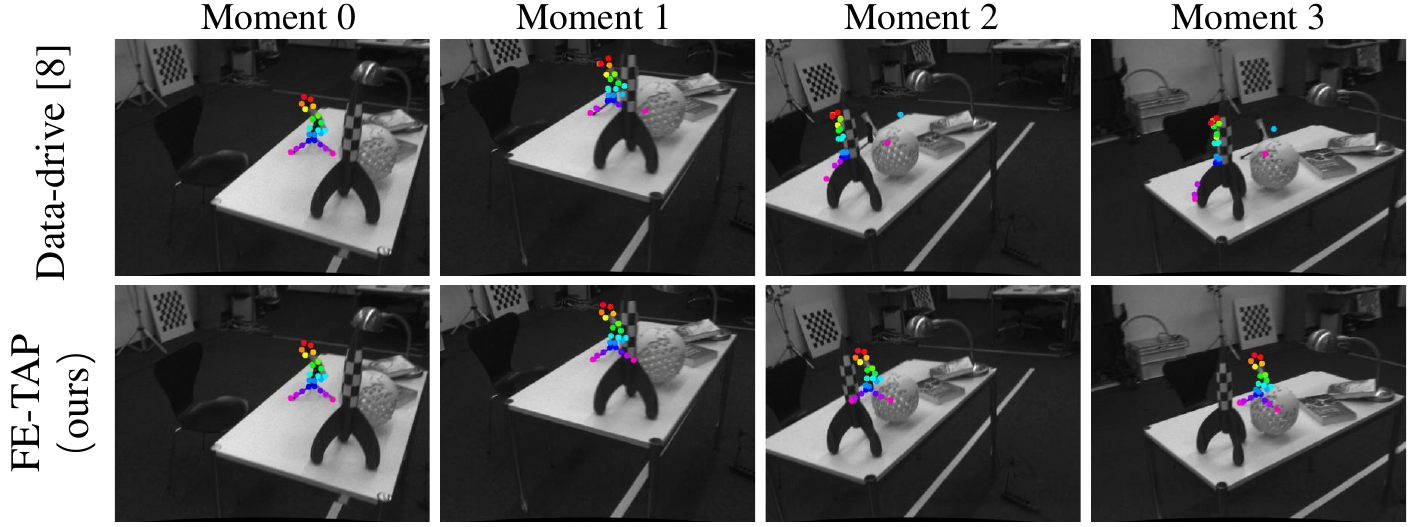}
    \caption{The comparison of our method and data-driven~\cite{conf_cvpr_MessikommerFG023} under occlusions}
    \label{fig:result_occupy}
\end{figure}

Due to the simultaneous optimization of target point trajectories within a sliding window, our model can handle occlusions within the window through attention mechanisms. This allows for temporal associations and utilizes surrounding spatial information to detect occluded points, as illustrated in  \cref{fig:result_occupy}. The first row shows the SOTA tracking method, Data-driven, while the second row presents our method. It is evident that, even in the presence of occlusions, our method maintains accurate tracking of target points.

\subsection{Ablation Study}
\begin{table}[t]
\centering
\caption{Ablation study on the setup of proposed FE-TAP}
\label{tab:ablation}
\begin{tabular}{lcc@{}}
\hline
 &Feature Age $\uparrow$& Expected Feature Age $\uparrow$  \\ \midrule
 (a).w/o accumulate & 0.696 & 0.657 \\
 (b).w/o time embed & 0.705 & 0.662 \\
 (c).w/o frame& 0.216 & 0.181 \\
 (d).w/o event& 0.627 & 0.572 \\ \midrule
 (e).FE-TAP(full)& 0.718  & 0.674   \\ \bottomrule
\end{tabular}
\vspace{-0.3cm}
\end{table}

In the ablation study, we set the total number of training steps to 100,000 and set the downsampling factor to 8 for efficiency, while keeping all other parameters the same as detailed in \cref{Implementation Details}. We tested our model on nine sequences from above datasets and reported the average FA and EFA across all test sequences. The results are shown in \cref{tab:ablation}.

\textbf{Impact of EvoFusion} To verify the effectiveness of our fusion module, we conducted experiments using a fixed time window for event data collection, where a simple convolutional network for fusion. The tracking results, presented in row (a) of \cref{tab:ablation}, show a notable reduction in FA and EFA compared to the full model. 
This suggests that while a simple convolutional network may struggle with precise image-event alignment, it can still effectively simulate image generation assisted by event streams.

\textbf{Impact of time embed token} We demonstrate the effectiveness of our enhanced trajectory iterative optimization model by removing the temporal information encoding from the fused features in the transformer model, as shown in \cref{tab:ablation} (b). The results indicate that incorporating temporal information significantly aids the transformer model in accurately identifying 
target point trajectories.

\textbf{Impact of Input Modalities} To validate the effectiveness of fusing images and events for point tracking, we tested our methods with only events and only images, excluding the other components mentioned above, see \cref{tab:ablation} (c-d). Due to the low resolution of the dataset and the downsampling factor set to 8, the model trained on synthetic events cannot be directly applied to real-world datasets. The image-only method resulted in a lower trajectory update frequency, while the absence of event information led to significant performance degradation in high-speed scenarios.

\subsection{Results in Driving Scenarios}

\begin{figure}[t!]
    \vspace{0.15cm}
    \centering
    \includegraphics[width=1\linewidth]{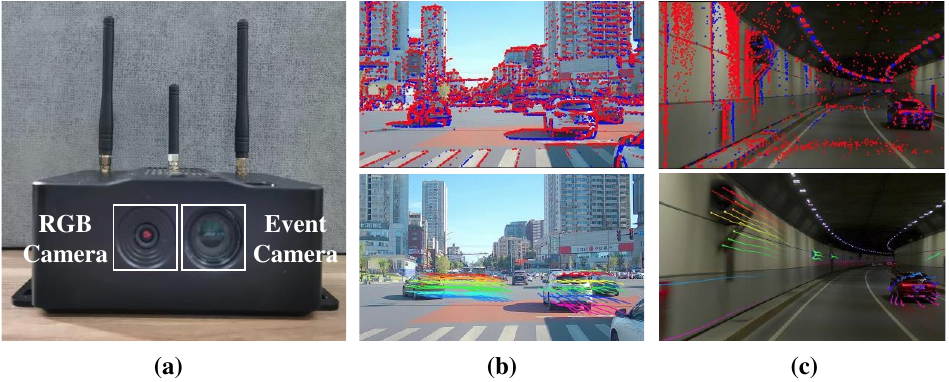}
    \vspace{-0.6cm}
    \caption{(a) Custom-designed image-event synchronization device; We validated the performance of our tracker in real-world driving scenarios, including urban roads (b) and tunnel (c) environments.}
    \vspace{-0.3cm}
    \label{fig:event-rgb}
\end{figure}

Since the datasets used previously only contain static indoor scenes with low resolution, we aimed to test our method's robustness in more complex environments with moving objects. To this end, we collected a real-world driving dataset using our custom-designed image-event synchronization device, as shown in  \cref{fig:event-rgb} (a). Our synchronization device consists of a sensing SG2-AR0231C camera with a resolution of 1980 $\times$ 1080 at 20Hz and a PROPHESEE EVK4 event camera with a resolution of 1280 $\times$ 720. Temporal synchronization is achieved through hardware, and spatial calibration is performed by converting the event stream into image frames. The qualitative tracking results are visualized in \cref{fig:event-rgb} (b-c). In these subfigures, the top row illustrates the mapping of images onto event data, while the bottom row presents the tracked point trajectories. Tracking was conducted on target points located on vehicles in two different motion states, see \cref{fig:event-rgb} (b), as well as on moving vehicles and stationary objects inside a tunnel, see \cref{fig:event-rgb} (c). The results demonstrate that our method maintains robust tracking performance even in such complex driving conditions.

\section{CONCLUSIONS}

In this paper,  we proposed FE-TAP, the first data-driven tracker designed for arbitrary points that integrates both image frames and events. We designed the EvoFusion module from a novel perspective to fuse images and events at high frame rate, thus avoiding the complex and error-prone alignment of images and events required in previous methods. Then, we proposed an Iterative Refinement module, which encodes the fused information into tokens to optimize and generate smoother and more accurate trajectories. Additionally, Our tracker outperforms state-of-the-art methods on two public datasets, and we verified FE-TAP's performance in real-world driving scenarios using our custom-designed image-event synchronization device. Future work will focus on improving the real-time capability of our model.










\bibliographystyle{IEEEtran}
\bibliography{root}

\end{document}